# MultiBanAbs: A Comprehensive Multi-Domain Bangla Abstractive Text Summarization Dataset

Md. Tanzim Ferdous[1][0009−0008−3426−5580], Naeem Ahsan Chowdhury[1][0009−0000−1652−6865], and Prithwiraj Bhattacharjee[1][0000−0001−9300−9351]

Department of Computer Science & Engineering,
Leading University, Sylhet-3112, Bangladesh
{cse_1912020112, naeemahsan_cse, prithwiraj_cse}@lus.ac.bd

**Abstract.** This study developed a new Bangla abstractive summarization dataset to generate concise summaries of Bangla articles from diverse sources. Most existing studies in this field have concentrated on news articles, where journalists usually follow a fixed writing style. While such approaches are effective in limited contexts, they often fail to adapt to the varied nature of real-world Bangla texts. In today's digital era, a massive amount of Bangla content is continuously produced across blogs, newspapers, and social media. This creates a pressing need for summarization systems that can reduce information overload and help readers understand content more quickly. To address this challenge, we developed a dataset of over 54,000 Bangla articles and summaries collected from multiple sources, including blogs such as Cinegolpo[1] and newspapers such as Samakal[2] and The Business Standard[3]. Unlike single-domain resources, our dataset spans multiple domains and writing styles. It offers greater adaptability and practical relevance. To establish strong baselines, we trained and evaluated this dataset using several deep learning and transfer learning models, including LSTM, BanglaT5-small, and MTS-small. The results highlight its potential as a benchmark for future research in Bangla natural language processing. This dataset provides a solid foundation for building robust summarization systems and helps expand NLP resources for low-resource languages.

## 1 Introduction

Abstractive text summarization in Bangla has been the subject of many previous works, and summaries of news articles are typically composed by journalists according to traditional reporting conventions. The current research proposes an approach developed on a more diversified dataset that includes writers and contributors beyond journalists, such as bloggers, content creators, and general users. This dataset consolidates a substantial volume of information from a wide range of sources. Given the extensive availability and long-term archives maintained by newspaper websites, a large number of articles have been collected from The Business Standard (around 12,000) and Samakal (about 42,000). Owing to public access limitations and the restricted archiving range of Bangla blog sites, only about 700 posts have been gathered from Cinegolpo. Additionally, transformer-based models such as BanglaT5-small and MT5-small have been applied to the abstractive summarization task. These models are based on the sequence-to-sequence architecture of transformers, enabling effective capture of contextual relationships within Bangla text. Both BanglaT5-small and MT5-small have been fine-tuned on the constructed dataset to generate high-quality abstractive summaries, demonstrating strong performance compared to traditional recurrent approaches. Their application further highlights the adaptability and advancement of neural architectures in Bangla text summarization. Model evaluation has been performed using standard quantitative metrics, including ROUGE and BLEU, which measure the quality of generated summaries by assessing their overlap with reference summaries. These metrics quantify performance in terms of alignment between predicted and ground-truth summaries and have been widely adopted in natural language processing research. To sum up, the contributions are as follows:

- The first multi-domain Bangla text summarization dataset is introduced. It captures diverse writing styles and patterns.
- The dataset is the largest multi-domain collection to date. It has 54,620 articles and their summaries from three different sources.
- Strong baselines are built using deep learning models. The results are comparable to or better than state-of-the-art models.

---

[1] https://cinegolpo.com/
[2] https://samakal.com/
[3] https://www.tbsnews.net/bangla/

## 2   Related Work

The study of Bangla text summarization has evolved significantly over the past decades, with early research primarily focusing on extractive techniques and limited datasets. One of the pioneering efforts by Islam et al. [1] introduced "Bhasa," a search engine and summarizer for Unicode Bangla text. This approach integrated modules such as tokenization, keyword search, and summary generation and represented one of the first attempts at combining search engine capabilities with text summarization for Bangla. Uddin et al. [2] adapted well-known English text summarization methods, such as location, cue, and phrase frequency, to the Bangla context, thus paving the way for further research on heuristic extractive approaches. Sarkar et al. [3] revisited extraction-based Bangla summarization with a three-step methodology. Subsequent refinements by Sarkar et al. [4] introduced more nuanced metrics, including word frequency similarity and sentence length, to enhance summary quality. Efat et al. [5] utilized sentence scoring within a theme-focused document framework, offering effective summarization for single-topic texts. The first notable move towards abstractive summarization for the Bangla language was made by Kallimani et al. [6], who employed class-based templates and attribute-based information extraction rules to generate summaries, recognizing the infancy of abstractive techniques in Bangla at the time. Multiple document summarization was first explored by Uddin et al. [7], with keyphrase-based summarization outperforming keyword-based methods in conveying document meaning. Heuristic and rule-based methods followed, such as those by Abujar et al. [8] and Ghosh et al. [9], covering multiple linguistic attributes. More recent efforts integrated mathematical formalisms with grammatical rules to enhance Bengali text summarization robustness by Sikder et al. [10]. Newer contributions since 2023 have focused on dataset creation and neural abstractive techniques. For instance, Khan et al. [11] introduced BanglaCHQ-Summ, the first publicly available abstractive summarization dataset specifically for Bangla medical consumer health questions, comprising 2,350 question-summary pairs and benchmarking state-of-the-art multilingual models. Additionally, Miazee et al. [12] proposed neural network architectures tailored for Bangla abstractive summarization, focusing on stability and efficiency to generate concise paragraph-level summaries. Other recent works propose unsupervised graph-based abstractive systems and comprehensive reviews of emerging Bengali summarization techniques, reflecting growing interest and innovation in dataset development and abstractive methodologies. Based on the progression of prior studies in Bangla text summarization, in this study, a large and diverse dataset is now being made publicly available to support further research and development in this field.

## 3  Dataset

Most Bangla text summarization research has used datasets from newspapers, where articles are mostly written by journalists and follow a uniform style. The BWSD [13] contains 1,100 web-sourced Bangla news articles with summaries, and MASBA [14] offers multi-level summaries of Bangla news articles. These datasets lack diversity in language, style, and vocabulary. To address this, this study builds a more varied dataset that combines newspapers, blogs, and business sources. It captures richer Bangla usage and enables models to generalize better across different text types.

### 3.1  Dataset Collection

About 54,620 articles have been collected from three different sources to capture diverse writing styles. Samakal is a major Bangla newspaper where professional journalists write formal news articles. Cinegolpo is a blogging platform with informal, story-like content on movies, series, and dramas. The Business Standard provides financial, business, and economic content written by professionals. To collect the data efficiently, web crawlers have been built for each source. The crawlers found article links, extracted text and summaries, removed ads and incomplete entries, and stored the cleaned data in a uniform format. The final dataset includes 41,675 articles from Samakal, 12,255 from The Business Standard, and 690 from Cinegolpo. By combining these sources, the dataset captures a wide range of linguistic patterns and writing styles. This makes it large, diverse, and well-structured. It also makes it suitable for preprocessing, model training, and testing summarization systems. The following table 1 mentions the summary of articles collected from different sources for the dataset.

Table 1: Summary of articles collected from different sources for the dataset.

| Source | Content Type | Number of Data Points |
|---|---|---|
| Samakal | News articles | 41,675 |
| The Business Standard | Financial, business, and economy contents | 12,255 |
| Cinegolpo | Blogs on movies, series, dramas | 690 |
| **Total** | | **54,620** |

### 3.2 Dataset Preprocessing

After collecting the raw dataset, we performed extensive preprocessing to clean and structure it for model training. The dataset contains 54,620 article–summary pairs with varying lengths. Articles range from 63 to 1,052 words, with an average of 262.21 words and a median of 210 words. The standard deviation is 164.59, showing high variation in article sizes. About 8.34 percent of articles (4,556 samples) exceed 512 words, which is important for sequence-to-sequence models with fixed input sizes. The article length distribution shows the 25th percentile at 153 words, the 50th at 210 words, the 75th at 307 words, and the 95th at 634 words. For summaries, lengths varied from a single word to a maximum of 63 words (489 characters). The average summary length was 30.41 words, with a median of 29 words and a standard deviation of 11.55 words. Only 25 summaries (0.05 percent) contained fewer than 10 words, showing that most summaries provided enough content. The 25th, 50th, 75th, and 95th percentiles were 21, 29, 38, and 52 words, respectively. On average, articles were 9.61 times longer than their summaries, with ratios ranging from 1.45 to 171.0, highlighting the compression challenge in abstractive summarization. The dataset included 346,869 unique tokens, with 50.3 percent classified as rare words that made up only 1.09 percent of all tokens. The most common tokens were function words and frequent verbs, such as **"ও"** (189,247), **"করে"** (133,663), **"এ"** (116,981), **"থেকে"** (106,175), and **"করা"** (101,189). These preprocessing steps ensured the dataset was clean, structured, and well-understood for modeling. Figure 1 depicts the horizontal box plots showing the distribution of article and summary lengths.

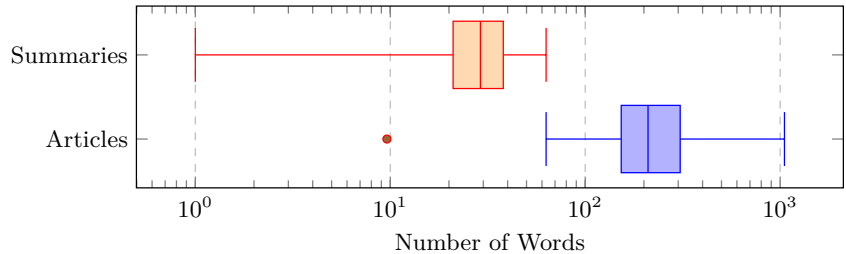

Fig. 1: Box plots showing the distribution of article and summary lengths.

The x-axis is logarithmic to visualize both articles and summaries clearly. A red dot shows the average article-to-summary length ratio.

### 3.3 Dataset Comaparison

To show the value of the proposed dataset, a comparison of it with other Bengali text summarization datasets has been shown. Table 2 lists existing datasets and the proposed dataset with 54,620 articles and summary pairs. BNLPC [15] has only 200 articles with three summaries each. XL-Sum [16] contains 10,126 articles with one summary per article. BANSData [17] has 19,096 articles and summary pairs focused on news. MASBA [14] has about 54,000 articles with three types of summaries per article. The proposed dataset has 54,620 articles with one summary each. It is the largest single-summary multi-domain dataset and covers more content than XL-Sum [16] and BANSData [17]. It provides a consistent one-to-one mapping, which is useful for training baseline models. This shows that our dataset is larger and more diverse than existing resources and is valuable for developing Bangla summarization systems. We have made our "MultiBanAbs" publicly available at Kaggle[4].

Table 2: Comparison of Existing Datasets with our Proposed Dataset

| Dataset Name | Source Type | Number of Articles | Number of Summaries |
|---|---|---|---|
| BANSData | News Article | 19,096 | 19,096 |
| BNLPC Dataset | Wikipedia, News Article | 200 | 600 |
| XL-Sum | News Article | 10,126 | 10,126 |
| BWSD | News Article, Facebook, Blogs | 1,100 | 1,100 |
| MASBA | News Article | 160,333 | 3 summaries per article |
| BUSUM-BNLP | News Article | 19,000 | 19,000 |
| Proposed Dataset | **Cinema Blogs, News Article, Business Article** | 54,620 | 54,620 |

## 4 Baseline Evaluation

Previous research on Bangla text summarization explored various models. Early works like Singha et al. [18] used LSTM-based recurrent networks and attention mechanisms on datasets like BANSData. More recent studies applied transformer-based models such as mT5 [19] and BanglaBERT [20]. It leverages large-scale pretraining for fluent and accurate summaries. Miazee et al. [12] proposed stable neural summarization frameworks. In 2023, a ranking-based approach showed that combining pretrained summarizers with ranking algorithms improves output quality. In this study, mT5 and BanglaT5 [21] were chosen as baselines. mT5 provides multilingual transfer learning. BanglaT5, trained specifically on Bangla corpora, ensures strong linguistic coverage. These models form robust baselines for evaluating the proposed dataset and future Bangla summarization research. In Figure 2 Methodological Flow of the Baseline Models is shown.

---

[4] https://www.kaggle.com/datasets/naeem711chowdhury/multibanabs

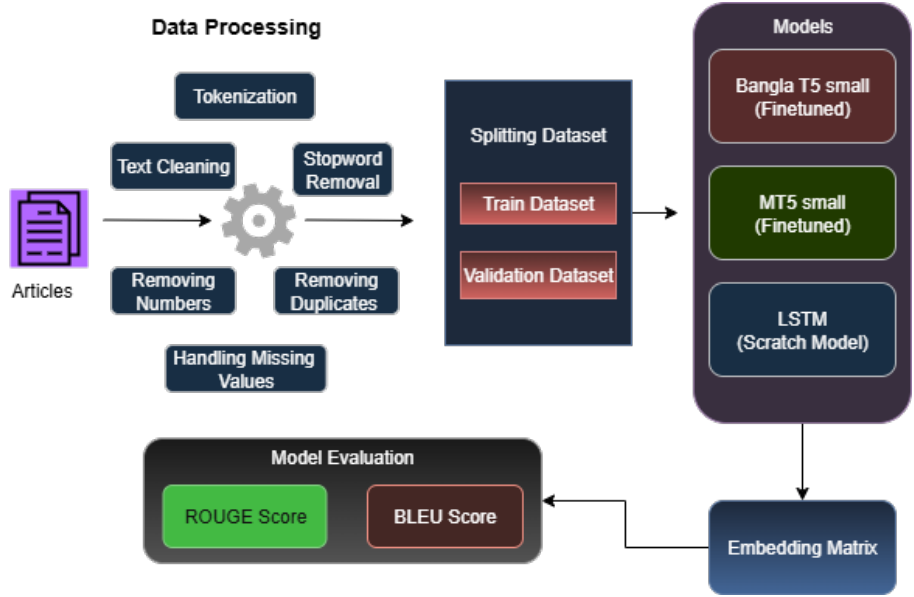

Fig. 2: Methodological Flow of the Baseline Models

- **BanglaT5-small as a Baseline Model:** The BanglaT5-small model is a transformer encoder and decoder. It has six encoder layers and six decoder layers. There are twelve transformer blocks in total. Each layer has a hidden size of 512 and a feed-forward size of 2048. The model has about 60.5 million trainable parameters. The encoder uses self-attention and feed-forward layers. The decoder uses self-attention and cross-attention to generate summaries. The model provides around 30,720 neurons per token. It is efficient and expressive. It works well for Bangla abstractive summarization with limited resources.
- **mT5-small as a Baseline Model:** The mT5-small model was used as another baseline. It has strong multilingual pretraining. It works well for low-resource languages like Bangla. The model has a transformer encoder and decoder. There are eight encoder layers and eight decoder layers. It has sixteen transformer blocks in total. Each block has a hidden size of 512 and a feed-forward size of 1024. The model uses GELU activation functions. It has about 300 million trainable parameters. It provides around 24,576 neurons per token. The model can capture syntactic and semantic relationships. It can transfer knowledge from other languages. It is robust for Bangla tasks and a strong baseline for summarization.
- **3-Layer LSTM as a Baseline Model:** The 3-layer LSTM model was used as a baseline. The model has three stacked LSTM layers. An embedding layer converts input tokens into vectors. The vectors pass through the LSTM layers sequentially. The model has about 16.1 million trainable parameters. The first and third layers have 481,200 parameters each. The second layer has 721,200 parameters. The model learns higher-level patterns from the input. It was trained with supervised learning. Generated summaries were compared to reference summaries. The weights were updated with backpropagation. This LSTM is simpler than transformers, but captures sequential information well. It is a strong baseline for Bangla abstractive summarization.

## 5  Result

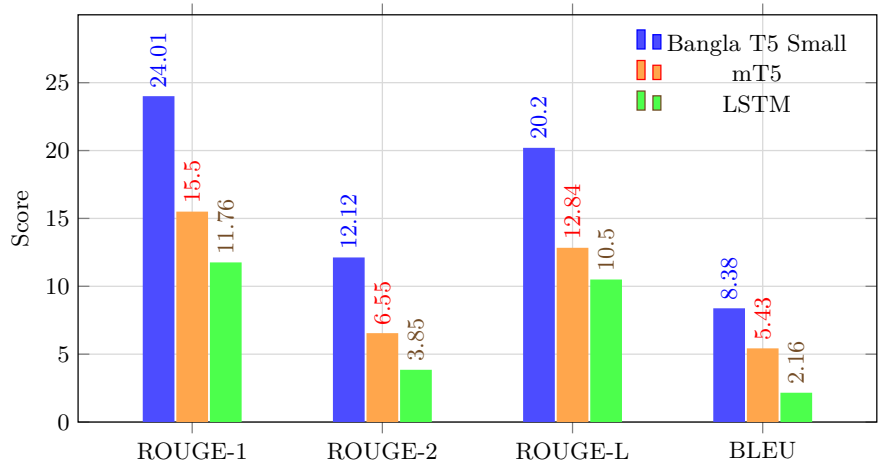

Fig. 3: Comparison of ROUGE and BLEU scores of Three Baselines

In this study, to evaluate the performance of our baseline systems, a system-oriented evaluation approach is employed, where the program is given both the actual and predicted summaries as input, and it generates a score comparing how much the predicted summary deviates from the actual summary. ROUGE-1, ROUGE-2, and ROUGE-L scores have been calculated. Barbella et al. [22] discuss the common use of ROUGE-1, ROUGE-2, and ROUGE-L for different levels of text granularity in summarization evaluation. Precision and Recall for the ROUGE technique were first calculated. Using these two measurements, the F1 score is calculated and takes the average across multiple test instances. This method of evaluation, on ROUGE and BLEU scoring are well-established method for evaluating the quality of summaries. ROUGE-1 measures unigram overlap, ROUGE-2 captures bigram overlap, and ROUGE-L computes the longest common subsequence between predicted and reference summaries.

Table 3: Reference and Predicted Summaries for Different Models

| Models | Reference Summary | Predicted Summary |
|---|---|---|
| **Bangla T5-small** | চট্টগ্রাম ট্রাক হামলা মামলার রায় ঘোষণার দুই দিন আগে চট্টগ্রাম আদালত ভবনে অগ্নিসংযোগের ঘটনায় সাধারণ ডায়েরি দায়ের হয়েছে।<br>**English:** A general diary GD has been filed at the police station over the fire incident at the Chattogram court building, which occurred two days before the verdict in the sensational truck arms case. | চট্টগ্রাম ট্রাক হামলা মামলার রায় ঘোষণার সময় আদালত ভবনের স্টোররুম ও রেকর্ড রুমে অগ্নিসংযোগের ঘটনায় চট্টগ্রাম আদালতের নাজির আবুল কালাম আজাদ সাধারণ ডায়েরি দায়ের করেছেন।<br>**English:** In connection with the incident of setting fire to the veranda of the storeroom and record room during the verdict announcement of the truck arms case in Chattogram, Abul Kalam Azad, the Nazir of the Chattogram CMM Court, has filed a general diary. |
| **mT5-small** | গাজীপুরের কালিয়াকৈরে সড়ক দুর্ঘটনায় পুলিশের এক কনস্টেবল নিহত হয়েছেন। মঙ্গলবার সন্ধ্যায় কালিয়াকৈর উপজেলার চন্দ্রা বাইপাস সড়কের পাশে দুর্ঘটনাটি ঘটে।<br>**English:** In Kaliaganj of Gazipur, a police constable named Abdul Motalib was killed after being hit by a motorcycle. The accident occurred on Tuesday evening in the area adjacent to the Charsindur Bridge on the Gazipur Kaliaganj Charsindur Bypass Road in Moktarpur Union of the upazila. | রাজধানীর কালিয়াকৈর উপজেলার চন্দ্রা বাইপাস সড়কের পাশে এক মোটরসাইকেল চালকের বেপরোয়া গাড়ি চালানোর কারণে কনস্টেবল মতালিব আহত হন।<br>**English:** In the area adjacent to the Charsindur Bridge in Kaliaganj upazila of the capital, Constable Motalib was injured when he was hit by a motorcycle that was being driven recklessly. |
| **LSTM** | স্বাস্থ্য পরীক্ষা শেষে বিএনপি চেয়ারপারসন খালেদা জিয়া ফিরোজায় ফিরেছেন। সোমবার বিকেলে তিনি এভারকেয়ার হাসপাতাল থেকে বাসায় ফেরেন।<br>**English:** After a health check-up, BNP Chairperson Khaleda Zia has returned to her Gulshan residence, Firoza. This Monday afternoon, she went home from her residence to Evercare Hospital, underwent several tests, and returned home. | বিএনপি চেয়ারপারসন খালেদা জিয়ার চিকিৎসার জন্য তাকে দেখতে গেছেন খালেদা জিয়া।<br>**English:** For the treatment of BNP Chairperson Khaleda Zia, the former Khaleda Zia, the court has met Khaleda Zia. |

As illustrated in Figure 3, the bar diagram shows the comparison of ROUGE and BLEU scores across our three baseline models: Bangla T5 Small, mT5 Small, and LSTM. The results reveal significant performance differences among the architectures. The Bangla T5 Small model demonstrated superior performance across all evaluation metrics, achieving scores of 24.01 for ROUGE-1, 12.12 for ROUGE-2, 20.2 for ROUGE-L, and 8.38 for BLEU. The mT5 Small model showed moderate performance with scores of 15.5 for ROUGE-1, 6.55 for ROUGE-2, 12.84 for ROUGE-L, and 5.43 for BLEU. In contrast, the LSTM baseline exhibited considerably lower performance, obtaining scores of 11.76 for ROUGE-1, 3.85 for ROUGE-2, 10.5 for ROUGE-L, and 2.16 for BLEU. The substantial performance gap between the transformer-based models and the traditional LSTM architecture underscores the effectiveness of modern attention-based mechanisms for Bengali abstractive summarization. The Bangla T5 Small model consistently outperformed both mT5 Small and LSTM across all metrics, demonstrating the significant advantage of language-specific pre-training for Bengali text summarization. Our final results are considered significant by standard evaluation criteria. Table 3 illustrates the generated examples of the predicted summary by the three baseline models.

## 6  Conclusion

This work presents MultiBanAbs, the largest multi-domain Bangla abstractive summarization dataset to date. It contains 54,620 article-summary pairs across diverse sources such as news, business, and cinema blogs. Establishment of strong baselines with BanglaT5-small, mT5-small, and a 3-layer LSTM has been shown. BanglaT5-small consistently outperformed others with significant gains on ROUGE and BLEU metrics. These results highlight the value of language-specific pre-training for Bangla. To further promote transparency, reproducibility, and community-driven development, we make the dataset publicly available under a permissive open license. The dataset will be accessible for academic research and non-commercial purposes, with appropriate citation requirements. We also intend to release comprehensive documentation detailing the dataset structure, preprocessing procedures, and access instructions through an online repository. This planned release aims to ensure ethical utilization of the data and encourage future research, extensions, and benchmarking studies within the low-resource language summarization domain. By releasing MultiBanAbs, a rich resource has been provided to the community for advancing abstractive summarization research in low-resource languages. Future work includes extending domains and applying state-of-the-art transformer-based approaches for improved semantic and factual quality.